\documentclass[runningheads]{llncs}
\usepackage{graphicx}
\usepackage[misc]{ifsym}
\usepackage{todonotes}
\usepackage{blindtext,letltxmacro,xcolor,xparse}
\usepackage{lipsum}
\usepackage{siunitx}

\usepackage{caption}
\usepackage{subcaption}

\LetLtxMacro{\blindtextblindtext}{\blindtext}
\LetLtxMacro{\blindtextBlindtext}{\Blindtext}

\RenewDocumentCommand{\blindtext}{O{\value{blindtext}}}{%
  \begingroup\color{gray}\blindtextblindtext[#1]\endgroup
}
\RenewDocumentCommand{\Blindtext}{O{\value{blindtext}}O{\value{Blindtext}}}{%
  \begingroup\color{gray}\blindtextBlindtext[#1][#2]\endgroup
}

\begin{document}
\title{Fine-tuning Is a Surprisingly Effective Domain Adaptation Baseline in Handwriting Recognition}
\titlerunning{Handwriting Recognition Domain Adaptation Baseline}
%

\author{Jan Kohút (\Letter) \orcidID{0000-0003-0774-8903} \and
Michal Hradiš\orcidID{0000-0002-6364-129X}}

\authorrunning{J. Kohút et al.}
%
\institute{Faculty of Information Technology, Brno University of Technology, Brno, Czech~Republic \\
\email{ikohut@fit.vutbr.cz}, \email{ihradis@fit.vutbr.cz}}
\maketitle              
\begin{abstract}
In many machine learning tasks, a large general dataset and a small specialized dataset are available. 
In such situations, various domain adaptation methods can be used to adapt a general model to the target dataset. 
We show that in the case of neural networks trained for handwriting recognition using CTC, simple fine-tuning with data augmentation works surprisingly well in such scenarios and that it is resistant to overfitting even for very small target domain datasets. 
We evaluated the behavior of fine-tuning with respect to augmentation, training data size, and quality of the pre-trained network, both in writer-dependent and writer-independent settings. 
On a large real-world dataset, fine-tuning on new writers provided an average relative CER improvement of 25\,\% for 16 text lines and 50\,\% for 256 text lines.
\keywords{Handwritten text recognition  \and OCR \and Data augmentation \and Fine-tuning.}
\end{abstract}


\section{Introduction}

In handwriting recognition, an OCR trained on a large and general dataset is often used to transcribe new writers. 
These writer-independent models provide good accuracy; however, when the writing style of the new writer differs from the general dataset, the transcription accuracy degrades and some form of domain adaptation may become necessary. 
In fact, we believe that some form of domain adaptation should be performed whenever a larger collection of consistent texts is to be transcribed.
Although unsupervised strategies may be used, a couple of text lines from the target collection can be manually transcribed with minimal effort while often providing superior accuracy improvement.

In this paper, we explore domain adaption of large convolutional-recurrent CTC neural networks~\cite{E2ENNCTC2015,AREMDLSTMNECESSARY2017,GatedConv2017,IMPROVINGCNNBLSTMCTC2018} from a large general dataset of mostly modern handwriting to specific documents written in various languages and scripts.
Specifically, we fine-tune a general  model to a small number of annotated text lines from a target document 
with practical strategies for early stopping.
We show that this simple approach is a surprisingly effective domain adaptation baseline, especially with suitable data augmentation, even for an extremely low amount of annotated target data. 
The proposed approach is stable, simple to implement, and provides consistent improvements in a wide range of situations.
In fact, the fine-tuning approach is used in our text recognition web application PERO OCR\footnote{https://pero-ocr.fit.vutbr.cz} with great user feedback.

The specific contributions of this paper are as follows: (1) study of CTC network domain adaptation by fine-tuning on small datasets (1-256 text lines); (2) evaluation of possible variation of the improvement for multiple target documents written in different scripts and styles; (3) hyperparameter selection strategies suitable for fine-tuning in realistic scenarios; (4) convergence and overfitting analysis on small target datasets; (5) proposal of effective data augmentations and their evaluation; (6) strong evidence that fine-tuning is effective also in writer-dependent scenario (fine-tuning to documents or writer from the training set); (7) new dataset of 19  manuscripts suitable for domain adaptation experiments in various European languages and scripts with at least 512 hand-transcribed lines each.

\section{Related Work}\label{sec:related_work}
Modern handwritten text recognition approaches are either based on Connectionist Temporal Classification (CTC)~\cite{CTC2006} or are full seq2seq models with an autoregressive decoder. 
CTC models~\cite{E2ENNCTC2015,AREMDLSTMNECESSARY2017,GatedConv2017,IMPROVINGCNNBLSTMCTC2018} are usually based on a stack of convolutional layers, followed by LSTM blocks~\cite{LSTM1997}. 
Older seq2seq architectures~\cite{Seq2Seq2019,Seq2SeqEff2018} use encoders with similar architectures and decoders composed of LSTM blocks which are usually enhanced by various attention mechanisms.
Lately, the recurrent layers were replaced by Transformers~\cite{AttentionIsAllYouNeed2017} blocks where information in a sequence is distributed purely by self-attention mechanism.
Text recognition Transformers~\cite{PayAttentionTransformer2020,BiDecodingTransformer2021,Rethinking2021,LightTransformer2022,TrOCR2023} similarly to other models use convolutional layers in the encoder.
Based on the available literature, the mentioned architectures provide comparable transcription accuracy~\cite{Seq2Seq2019,PayAttentionTransformer2020,Rethinking2021}, while some works indicate that seq2seq model may prove to be superior as larger datasets become available~\cite{BiDecodingTransformer2021}.

Similar to our approach, several works~\cite{Boosting2021,Improving2019,OCR4AllHWRMedieval2022,OCR4AllPrinted2021} explored domain adaptation of CTC-based models by fine-tuning. 
However, the experiments did not explore the limits of such an approach (e.g. fine-tuning to less than a dozen lines), did not explore possible  strategies for choosing hyperparameters, and did not explore the tendency of overfitting in these scenarios. 
Also, some of the findings and observed behaviors are not consistent (e.g. effect of data augmentation).

Aradillas et al.~\cite{Boosting2021} experimented with domain adaptation from  IAM~\cite{IAM2002} dataset to Washington~\cite{George2004} and Parzival~\cite{Parzival2009} datasets, and between different partitions of the READ dataset~\cite{ICFHR2018}. 
Their conclusions are that it is better to fine-tune the network than selected network layers and that geometric augmentation~\cite{HWRDataAugmentation2017} of the target domain degrades final accuracy.
This is contrary to our findings, where our data augmentation combined with fine-tuning brought substantial increases in accuracy.
Soullard et al.~\cite{Improving2019} also experimented with fine-tuning on the READ dataset.
Similar to us, they used cross-validation to estimate the optimal number of fine-tuning iterations.
They used random rotation and scaling as data augmentation for both source and target domains. 
They also experimented with writers-specific language models which further improved results.
Reul et al.~\cite{OCR4AllHWRMedieval2022} tested domain adaptation using fine-tuning on German medieval manuscripts in Gothic and Bastarda scripts.
They utilized data augmentation in the form of several binarization strategies, both for source model training and fine-tuning.
They used an ensemble of models combined with a voting strategy optimized with cross-validation. 
However, the stopping criterion of the fine-tuning was controlled by the testing datasets error.
They observed that the closer the source model data was to the target data, the better the results after fine-tuning.

Bhunia et al.~\cite{MetaHTR2021} approached domain adaptation to new writers as a meta-learning task, where the goal is to train a general model that can be effectively adapted to new domains with a single update and few words. 
They found that a single-shot adaptation of such a general model is superior to fine-tuning a model trained in a standard fashion.
However, the experiments were restricted to word-level IAM and RIMES~\cite{RIMES2009} datasets and 16 adaptation words images. 
Instead of training a general model, Kohut et al.~\cite{Towards2023} proposed a model with dedicated writer-dependent parameters which can handle multiple writers simultaneously.
While adapting to a new writer, optimizing a new set of writer-dependent parameters brought worse performance than fine-tuning all parameters.

As speech and handwritten text recognition are closely related, we also present a short overview of domain adaptation from this field.
Hank Liao~\cite{FT:SpeakerAdaptationOfContextDependentDeepNeuralNetworks2013}
explored how a simple neural acoustic model may be adapted to speakers by fine-tuning the input layer, the output layer, or the entire network. 
Adapting the input layer was better than adapting the output layer, adapting all layers was even better.
In order to overcome overfitting, some strategies~\cite{KL-Divergence2013,AdversarialSpeakerAdaptation2019} regularize the fine-tuning process by minimizing the divergence between the feature distributions of the original network and of the fine-tuned one, where the features might be taken from any layer. 
These approaches require evaluation of the original network while fine-tuning the new one. 
Dong Yu et al.~\cite{KL-Divergence2013} minimized the senone distributions divergence by adding the Kullback–Leibler term to the loss function, which is equivalent to constructing the fine-tuning ground truth as linear interpolating of the fine-tuned and original model senone distribution.
Meng et al.~\cite{AdversarialSpeakerAdaptation2019,SpeakerAdaptationForE2E2019} forced the distribution of hidden features to be close with an adversarial approach, which is equivalent to minimizing the Jensen–Shannon divergence.

In scenarios where no annotated data for the target domain are available, unsupervised approaches in the form of consistency regularization~\cite{MultimodalConsistencyRegularization2022} and pseudo-labeling~\cite{ATST2021,FixMatch2022,JapanPseudoLabel2020} may be used.
In scenarios where both annotated and unannotated data are available, supervised and unsupervised approaches may be combined to get the best out of both worlds.
For example, fine-tuning together with pseudo-labeling~\cite{PseudoLabelingFinetuning2020}.

\section{CzechHWR Dataset}\label{sec:dataset}

We collected a large dataset of mainly 19\textsuperscript{th} and 20\textsuperscript{th} century Czech handwritten documents which, in our opinion, is a realistic example of a general dataset for training writer-independent models.
The CzechHWR dataset was created from three main sources: documents processed by users of our text recognition web application PERO OCR, a collection of Czech letters transcribed by linguists~\cite{hladka_2013}, and Czech chronicles transcribed specifically for handwriting recognition.
From the OCR application, we collected 295k text lines manually corrected by the users (after reviewing one or two pages from each user). 
The documents are mostly written in Czech modern cursive script, although a marginal part is written in German Kurrent and in several medieval scripts.
The original sources are mainly military diaries, chronicles, letters, and notes.
The Czech letters~\cite{hladka_2013} consists of 2000 letters (87k text line annotations) from 20\textsuperscript{th} century, mostly handwritten in Czech modern cursive with a limited amount of typeset ones.
We manually annotated approximately 2 pages of 277 distinct Czech chronicles, resulting in 553 pages with 24k text lines. 

\begin{figure}[t]
    \centering
    \includegraphics[width=\linewidth, trim=8mm 95mm 65mm 8mm, clip]{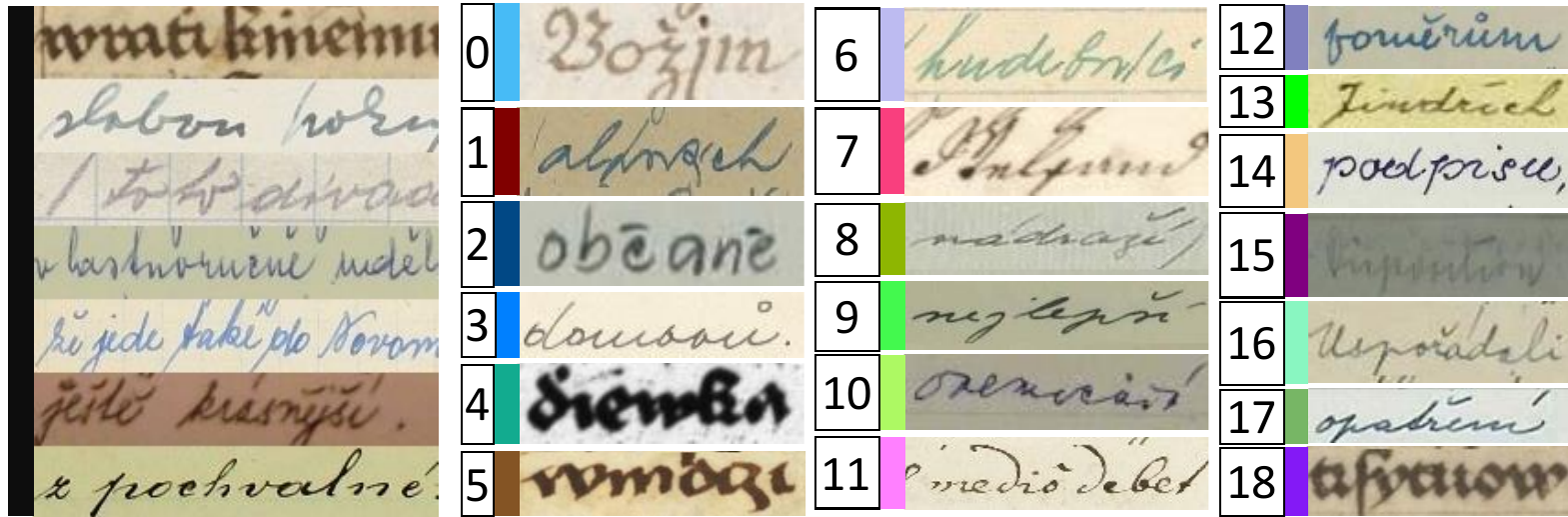}
    \caption{Black, samples from the large general source CzechHWR dataset. ID with Color, representative words of 19 target writers.}
    \label{fig:dataset:source_and_target}
\end{figure}

The final CzechHWR dataset contains 406k annotated text lines and our estimate of distinct writers is 4.5k.
The level of penmanship and readability differs, ranging from scribbles to calligraphy, although the tendency is towards fairly readable texts, see the left side of Figure~\ref{fig:dataset:source_and_target}.
The training (TRN) and testing (TST) subsets contain 379k and 5k lines.
Due to the fact that writers with a small number of total lines are not sufficiently represented in TST, we created $\mathrm{TST_{W}}$, which contains lines of all writers that have at least 20 lines in TRN.
Table~\ref{tab:dataset} shows the distribution of writers in the CzechHWR dataset according to the number of lines per writer with the respective amounts of lines for each subset.
\begin{table}[t]
\caption{The distribution of writers (NW) in the CzechHWR dataset according to the number of lines per writer (NWL) with the respective amounts of lines for each subset.}\label{tab:dataset}
\centering
{
\begin{tabular}{r | p{13mm} p{13mm} p{13mm} p{13mm} p{13mm} p{13mm} p{9mm} | l}
NWL &  1--19 & 20--49 & 50--99 & 100--199 & 200--499 & 500--999 & 1000-- & ALL \\
\hline\hline 
TRN & 13k & 79k & 82k & 43k & 24k & 16k & 122k & 379k \\
TST & 169 & 1k & 1.1k & 566 & 287 & 198 & 1.7k & 5k \\ 
$\mathrm{TST_{W}}$ & 0 & 4.5k & 6.2k & 3.2k & 2k & 1.1k & 5.4k &22.4k \\
\hline
NW & $\sim$1.1k & $\sim$2.3k & $\sim$1.2k & $\sim$322 & $\sim$79 & $\sim$21 & $\sim$54 & $\sim$5.1k \\
\end{tabular}}
\end{table}

We chose additional 19 writers from our PERO OCR web application as the small target datasets for fine-tuning\footnote{https://pero.fit.vutbr.cz/handwriting\_adaptation\_dataset}, each writer is represented by at least 512 lines, and each line is at least 30 characters long.
For each writer, an image of a representative word is shown in Figure~\ref{fig:dataset:source_and_target}, the colors match the colors in fine-tuning experiments graphs (Figure~\ref{fig:writer_finetuning} and Figure~\ref{fig:finetuning_curves}), and the IDs match the IDs in Table~\ref{tab:target_absolute_results}.
The scripts of these target writers range from some which are very similar to the majority of CzechHWR to some which are very different.



Our neural network architecture is similar to the state-of-the-art architectures for text recognition~\cite{E2ENNCTC2015,AREMDLSTMNECESSARY2017,GatedConv2017,IMPROVINGCNNBLSTMCTC2018} trained with CTC loss~\cite{CTC2006}.
It consists of a convolutional stage (CNN), inspired by the standard VGG arhitectures~\cite{VGG2014}, and a parallel bidirectional LSTM~\cite{LSTM1997} recurrent stage (RNN), which processes the input at multiple scales.

We trained the network with Adam~\cite{Adam2015} optimizer for 500k iterations up until convergence.
We used polynomial warmup of a third order to gradually increase the learning rate from 0 to \num{3e-4} in the first 10k iterations.
At iterations 200k and 400k, we used the warmup again, but the learning rate maximums were \num{0.7e-4} and \num{0.175e-4}. 
The batch size was set to 32 and we used the B1C1G1M1 augmentation (see Section~\ref{sec:data_augmentations}).
The system reached CER of 0.51\,\%, 2.17\,\%, 2.26\,\% on TRN, TST, and $\mathrm{TST_W}$ subsets respectively, and the CER on augmented TRN subset was 2.4\,\%.
The distribution of test CER on the small target datasets, had a mean of 5.17\,\%, a standard deviation of 4.82\,\%, a minimum of 0.62\,\%, and a maximum of 14.46\,\%

\subsubsection{Architecture details.}
The architecture is equivalent to our baseline TS-Net architecture~\cite{TS-Net2021}, a more detailed description together with a detailed diagram can be found in the referenced work. 
CNN is a sequence of 4 convolutional blocks, where each block has 2 convolutional layers with numbers of output channels set to 64, 128, 256, and 512, respectively.
All convolutional blocks except the last one are followed by a max pooling layer.
The CNN subsamples an input text line image by a factor of 4 in width.
The RNN consists of three parallel LSTM branches and one final LSTM layer.
The branches process scaled variants of the input with two LSTM layers, the scaling factors are 1, 0.5, and 0.25.
The outputs are upsampled back to the original dimension and their summation is processed by the final LSTM layer.
Each LSTM layer is bidirectional and has a hidden feature size of 256 for both directions.
The output of RNN is processed by a 1D convolutional layer with a kernel size of 3.

\section{Data augmentations}\label{sec:data_augmentations}

We chose various augmentations strategies to enlarge the amount of data artificially and to regularize the fine-tuning process.
We used combinations of four basic augmentations: NoiseBlurGamma, Color, Geometry, and Masking.
NoiseBlurGamma applies random motion blur, gauss noise, and gamma correction.
Color randomly changes brightness, contrast, saturation, and hue changes.
Geometry randomly adjusts text slant, horizontal scale, and vertical scale. 
Masking stands for random noise patch masking.
The height of a noise patch is the same as the height of text line images, the width is chosen randomly up to the width of approximately two letters, and multiple masking patches can be applied to a single text line image. 
The intuition behind noise masking is to strengthen the language modeling capability of the system.

\begin{table}[t]
\setlength{\tabcolsep}{4pt}
\caption{Our augmentations as combinations of four basic ones: NoiseBlurGamma (B), Color (C), Geometry (G), and Masking (M). 
The number of dots specifies the level of augmentation intensity (1, 2, 3).}\label{tab:augmentations}
\centering
{
\begin{tabular}{r | p{10pt} p{10pt} p{10pt} p{10pt} p{10pt} p{10pt} p{10pt} p{10pt} p{10pt} p{10pt} p{10pt} p{10pt} p{10pt} p{10pt}}
 & \rotatebox[origin=l]{90}{NONE} & \rotatebox[origin=l]{90}{B1} & \rotatebox[origin=l]{90}{B1C1} & \rotatebox[origin=l]{90}{B1G1} & \rotatebox[origin=l]{90}{B1C1G1} & \rotatebox[origin=l]{90}{B1C1G1M1} & \rotatebox[origin=l]{90}{B2C1G1M1} & \rotatebox[origin=l]{90}{B2C2G1M1} & \rotatebox[origin=l]{90}{B2C1G2M1} & \rotatebox[origin=l]{90}{B2C1G3M1} & \rotatebox[origin=l]{90}{B2C2G2M1} & \rotatebox[origin=l]{90}{B2C2G3M1} & \rotatebox[origin=l]{90}{B2C2G2M2} & \rotatebox[origin=l]{90}{B2C2G3M2} \\
\hline\hline 
BlurNoiseGamma & & $\bullet$ & $\bullet$ & $\bullet$ & $\bullet$ & $\bullet$ & $\bullet\bullet$ & $\bullet\bullet$ & $\bullet\bullet$ & $\bullet\bullet$ & $\bullet\bullet$ & $\bullet\bullet$ & $\bullet\bullet$ & $\bullet\bullet$ \\
Color & & & $\bullet$ & & $\bullet$ & $\bullet$ & $\bullet$ & $\bullet$$\bullet$ & $\bullet$ & $\bullet$ & $\bullet$$\bullet$ & $\bullet$$\bullet$ &  $\bullet$$\bullet$ & $\bullet$$\bullet$\\ 
Geometry & & & & $\bullet$ & $\bullet$ &  $\bullet$ & $\bullet$ & $\bullet$ & $\bullet$$\bullet$ & $\bullet$$\bullet$$\bullet$ & $\bullet$$\bullet$ & $\bullet$$\bullet$$\bullet$ & $\bullet$$\bullet$ & $\bullet$$\bullet$$\bullet$ \\
Masking  & & & & & & $\bullet$ & $\bullet$ & $\bullet$ & $\bullet$ & $\bullet$ & $\bullet$ & $\bullet$ & $\bullet$$\bullet$ & $\bullet$$\bullet$ \\

\end{tabular}
}
\end{table}
Table~\ref{tab:augmentations} shows the final augmentations in columns as combinations of the basic ones. 
If a basic augmentation is a part of the final one, the probability of applying it on the input is 0.2 for the NoiseBlurGamma, 0.333 for the Color, 0.66 for the Geometry, and 0.5 for the Masking, therefore all the augmentations allow the network to see the original text line images.
The number of dots specifies the level of augmentation intensity, the higher the number, the greater the range of randomness in the respective image operations.
There are two levels (1, 2) for NoiseBlurGamma (B), Color (C), and Masking (M) augmentations, and three levels (1, 2, 3) for Geometry (G) augmentation.
We refer to the final augmentations with abbreviations e.g. augmentation B2C1G3M1 is a combination of NoiseBlurGamma level~2, Color level~1, Geometry level~3, and Masking level~1.
\begin{figure}[t]
    \centering  \includegraphics[width=\linewidth, trim=5mm 5mm 5mm 5mm, clip]{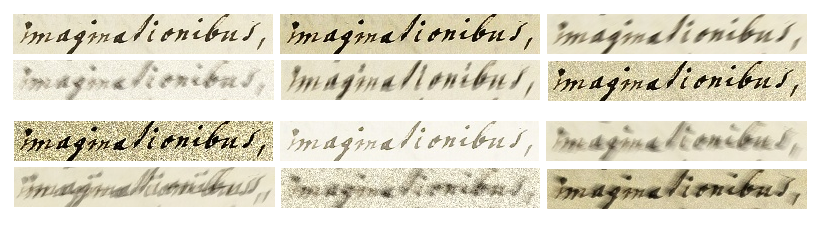}
    \caption{Augmented versions of top left text line image with NoiseBlurGamma augmentation. Intensity 1 is shown in the top section and intensity 2 in the bottom one. Only the extreme samples of the distributions are shown.} \label{fig:augmentations:noise_blur_gamma}
\end{figure}
\begin{figure}[t]
    \centering  \includegraphics[width=\linewidth, trim=5mm 5mm 5mm 5mm, clip]{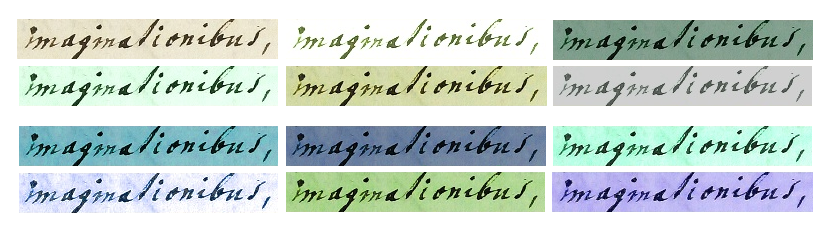}
    \caption{Augmented versions of top left text line image with Color augmentation. Intensity 1 is shown in the top section and intensity 2 in the bottom one. Only the extreme samples of the distributions are shown.} \label{fig:augmentations:color}
\end{figure}
\begin{figure}[t]
    \centering  \includegraphics[width=\linewidth, trim=5mm 5mm 5mm 5mm, clip]{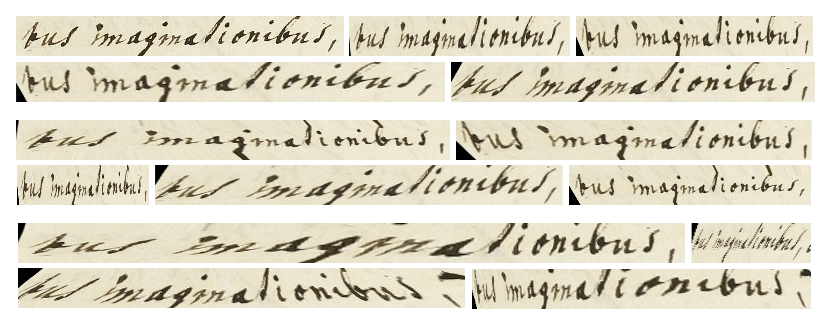}
    \caption{Augmented versions of top left text line image with Geometry augmentation. Intensity 1 is shown in the top section, intensity 2 in the middle one, and intensity 3 in the bottom one. Only the extreme samples of the distributions are shown.} \label{fig:augmentations:geometry}
\end{figure}
Figures~\ref{fig:augmentations:noise_blur_gamma},~\ref{fig:augmentations:color}, and~\ref{fig:augmentations:geometry} show augmented versions of the top left text line image with NoiseBlurGamma, Color, and Geometry, respectively.
For each level of augmentation intensity, there is a separate section of lines and only extreme samples are shown.

\section{Writer-independent Scenario}

In writer-independent scenario experiments, we fine-tuned the source baseline model trained on the CzechHWR dataset to the 19 target writers.
Experiments were based on writer fine-tuning runs.
A writer fine-tuning run consisted of drawing 512 random lines of the respective target writer and splitting them into 256 testing and 256 adaptation ones.
The adaptation lines were furthermore divided into 9 line clusters: 1, 2, 4, 8, 16, 32, 64, 128, and 256, where the numbers referred to the number of adaptation lines in them, and a smaller cluster was always a subset of all the larger ones.
The numbers of fine-tuning iterations were 200, 200, 400, 800, 1000, 1500, 2000, 2500, and 3000 for 1, 2, 4, 8, 16, 32, 64, 128, and 256 adaptation lines, respectively.
We run 10 fine-tuning runs for each writer resulting in total $19\times9\times10$ baseline model fine-tunings.
Additionally, we run $19\times5\times10$ 4-fold cross-validations for the line clusters 16, 32, 64, 128, and 256, as cross-validation on less than 16 lines is not reliable.

We estimated the optimal number of fine-tuning iterations, with different estimation strategies (ET).
The baseline estimation strategies were Last Iteration (L) and Oraculum (O).
Last Iteration (L) returned the last/maximum iteration.
Oraculum (O) returned the iteration of minimal CER on testing lines.
As there are no testing lines in practice, we experimented with estimation strategies based on 4-fold cross-validation computed on adaptation lines.
Minimum Iteration Average Across Chunks (A) smoothed each of 4 cross-validation test loss curves with window size 4, averaged the smoothed loss curves, and returned the iteration of the minimum loss.
Mean Minimum Iteration Per Chunk (M) smoothed each of 4 cross-validation test loss curves with window size 4, took the iteration of the minimum loss per each smoothed loss curve, and returned the mean of these iterations.
Max Minimum Iteration Per Chunk (X) estimated the optimal fine-tuning iteration in the same way as M, but at the end, instead of mean, returned the maximum. 
Note, as we tested every 20 iterations, the window size of 4 spanned across 80 iterations. 

We also experimented with a scenario, where there are multiple target writers with testing lines available and we want to assume a static number of fine-tuning iterations for a new target writer for which we do not have any testing lines.
The optimal static iteration was estimated on the writers with testing lines as the iteration of the minimum value of the writers' fine-tuning test curve.
The writers' fine-tuning test curve was calculated as an average of writer fine-tuning curves, which were normalized by their minimums.
Each writer's fine-tuning curve was calculated as an average of 10 fine-tuning test loss curves (10 fine-tuning runs), which were smoothed with a window size of 4. 
We refer to this estimation strategy as Static Iteration (S).
To compare it to others, we evaluate it in a 1 to N-1 manner, where N is the number of all target writers.

As the estimation based on loss often underestimated the number of optimal iterations (see Figure~\ref{fig:finetuning_curves}), we also experimented with simple modifications of X and S, denoted as $\mathrm{X_R}$ and $\mathrm{S_R}$, which multiplied the estimated iterations by a positive factor R of 1.5 and 3.
Estimations of these strategies were limited by the actual number of fine-tuning iterations.

\begin{figure}[t]
    \centering  \includegraphics[width=\linewidth, trim=0mm 0mm 0mm 0mm, clip]{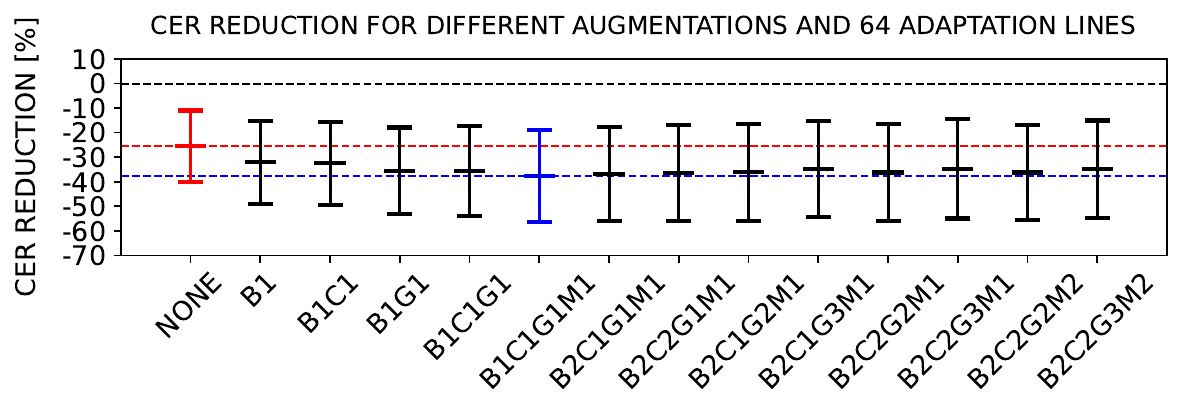}
    \caption{The performance of models fine-tuned with different augmentations expressed as a relative reduction of the baseline model test CER.
    The means and the standard deviations represent the target writer distribution.}
    \label{fig:augmentations:comparison}
\end{figure}
\subsubsection{Choosing augmentation for fine-tuning.}
Figure~\ref{fig:augmentations:comparison} compares the performance of models fine-tuned with different augmentations (see Section~\ref{sec:data_augmentations}) to the performance of the baseline model, which served as the starting point for the fine-tuning.
The comparison is expressed as a relative reduction of the baseline model test CER and it is given by:
\begin{equation}
\frac{F-B}{B}, 
\end{equation}
where B is the test CER of the baseline model and F is the test CER of its fine-tuned variant.
Due to the high number of augmentations, we run fine-tuning runs just with the line cluster 64.
$\mathrm{X}_{1.5}$ was used as the estimation strategy for choosing the fine-tuning iteration of minimal CER.
For each augmentation, the mean and the standard deviation of test CER reductions on all 19 target writers are shown.
\emph{In all experiments}, test CER reduction on a writer is the mean of test CER reductions across all 10 fine-tuning runs.
Fine-tuned models consistently outperformed the baseline model on most of the target writers and they did not worsen the accuracy on any.
NoiseBlurGamma, Geometry, and Masking augmentations improved the performance significantly, whereas Color augmentation had almost no effect.
The higher levels of augmentation intensity (2, 3), did not bring any essential variations in performance.
Even though the Geometry augmentations affected the writing style significantly (see Figure~\ref{fig:augmentations:geometry}), they consistently brought better performance for the fine-tuning.
On average, for line cluster 64, fine-tuning without any augmentation (NONE) reduced the baseline CER by 25\,\%, while combinations of all the basic augmentations reduced the CER by an additional 10\,\%.
Fine-tuning with data augmentations brought larger standard deviations across target writers. 
Furthermore, we only experiment with augmentations NONE and B1C1G1M1.

\begin{figure}[t]
    \centering  \includegraphics[width=\linewidth, trim=0mm 0mm 0mm 0mm, clip]{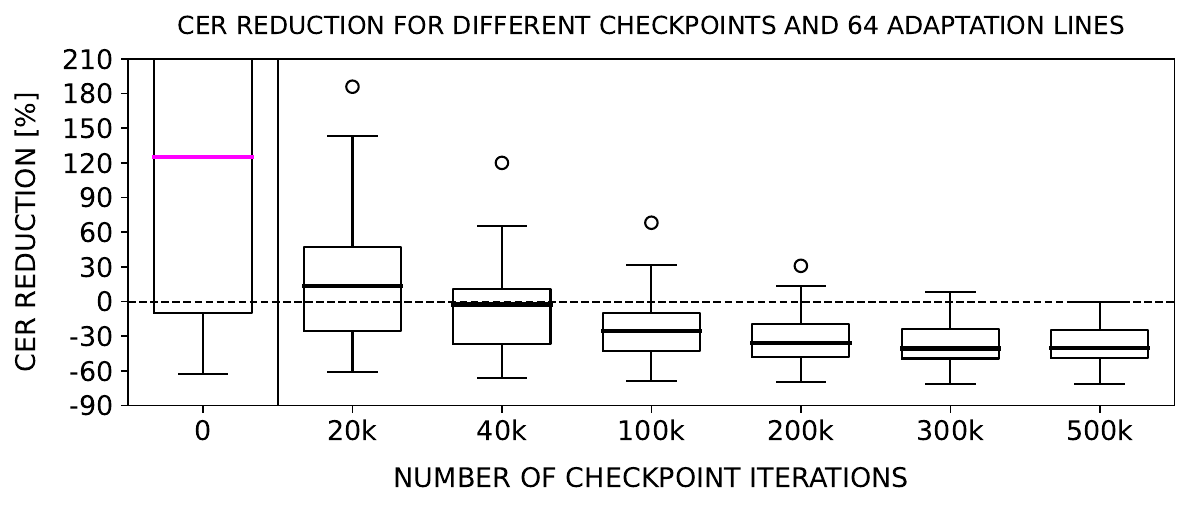}
    \caption{
    Fine-tuning of baseline models trained for different amounts of iterations on the CzechHWR dataset.
    The performance is expressed as a relative reduction of the fully-trained baseline model (500k) test CER. 
    The boxplots represent the target writer distribution.
    See the text for a description of the model fine-tuned from scratch (0).
    } \label{fig:checkpoints:comparison}
\end{figure}
\subsubsection{Pre-trained quality of the baseline model.}
Figure~\ref{fig:checkpoints:comparison} compares fine-tuning of baseline models trained for different amounts of iterations on the CzechHWR dataset.
The performance is expressed as a relative reduction of the fully-trained baseline model test CER (500k).
The boxplots represent the target writer distribution.
As with the previous experiment, we run the fine-tuning runs only for line cluster 64 and estimated the optimal fine-tuning iterations with $\mathrm{X}_{1.5}$.
The more well-trained the baseline model, the greater and more stable the performance across the target writers.

The architecture fine-tuned from scratch is almost identical to ours (described in Section~\ref{sec:dataset}), where the only essential difference is that the convolutional layers are initialized from VGG~\cite{VGG2014} architecture trained on ImageNet~\cite{deng2009imagenet}.
The fine-tuning was done on line cluster 256, for 10k iterations, and we estimated the number of optimal fine-tuning iterations with the Oraculum strategy (no cross-validation involved). 
In comparison to the well-trained baseline model, the performance was far worse for most of the writers, although there were exceptions among writers whose writing styles were not sufficiently represented in the CzechHWR dataset. 
Even though that VGG was trained on four times more lines and the Oraculum estimating strategy was used, it is surprising, that for writers with German Kurrent and Ghotic script, it evened out the fine-tuned well-trained baseline model.

\begin{figure*}[t]
    \centering
    \begin{subfigure}[t]{0.5\textwidth}
        \centering
        \includegraphics[width=\textwidth, trim=0mm 0mm 0mm 0mm, clip]{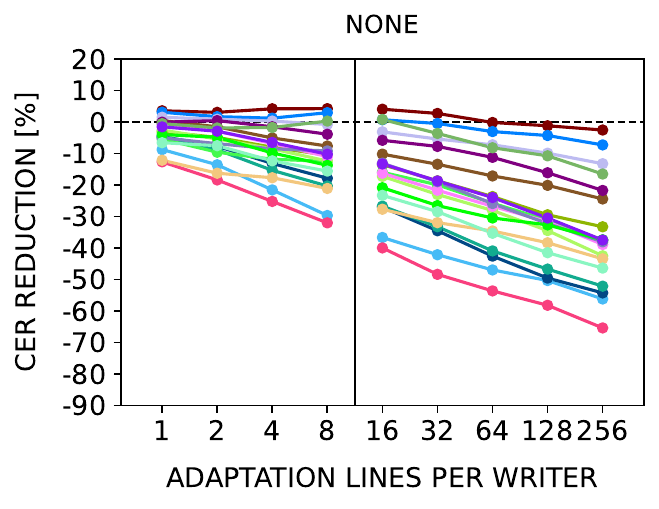}
    \end{subfigure}%
    ~
    \begin{subfigure}[t]{0.5\textwidth}
        \centering
        \includegraphics[width=\textwidth, trim=0mm 0mm 0mm 0mm, clip]{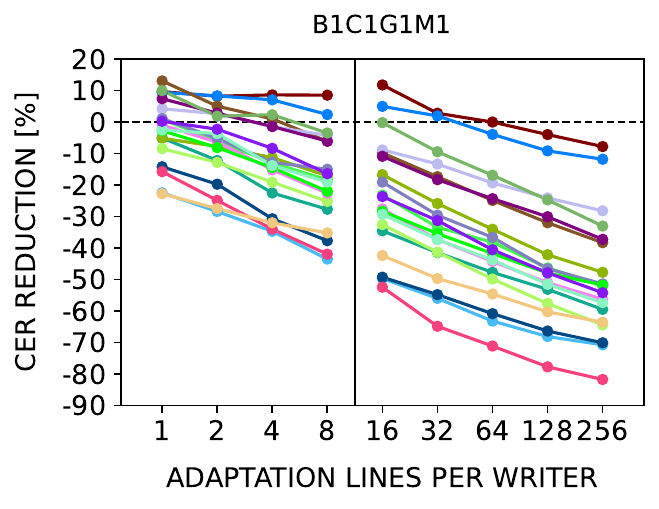}
    \end{subfigure}
    \caption{Relative reductions of the baseline model test CER on the target writers for complete fine-tuning runs, with and without augmentation.
            The estimation strategy for choosing the fine-tuning iteration with minimal test CER was $\mathrm{S}_{3}$ for 1--8 and $\mathrm{X}_{1.5}$ for 16--256 adaptation lines.}
    \label{fig:writer_finetuning}
\end{figure*}
\subsubsection{Fine-tuning runs.} Figure~\ref{fig:writer_finetuning} shows relative test CER reductions of the baseline model test CER on the target writers for complete fine-tuning runs, with and without augmentation.
The estimation strategy for choosing the fine-tuning iteration with minimal test CER was $\mathrm{S}_{3}$ for line clusters 1, 2, 4, 8 and $\mathrm{X}_{1.5}$ for line clusters 16, 32, 64, 128, and 256. 
Each writer is represented by a different color (see Figure~\ref{fig:dataset:source_and_target} for images of representative words).
Static iteration setup improved the performance even for 1 adaptation line, however, it overfitted two writers for all cluster lines.
Cross-validation setups improved the performance on all cluster lines, except for the same two writers in the case of line clusters 16 and 32.

Generally, the more adaptation lines, the greater the performance.
Fine-tuning with B1C1G1M1 augmentation consistently outperformed fine-tuning without any augmentation, although there is a higher risk of worsening the performance when fine-tuning with smaller amounts of lines.
The distribution across the target writers is Gaussian-like, while the augmentation shifts the mean, and stretches the standard deviation.
The largest CER reductions (up to 82\,\%) were achieved for distinct yet to some extent source-like writing scripts such as Kurrent or Czech block letters.
The average CER reductions (up to 60\,\%) were achieved for vastly different scripts such as Ghotic, and for harder-to-read source-like scripts.
For easy-to-read source-like scripts, smaller CER reductions (up to 10\,\%) were achieved for larger amounts of adaptation lines, whereas overfitting led to worse performance (up to 15\,\%) for smaller amounts of lines.
\begin{table}[t]
\caption{Test CER (in \%) of the baseline model (0) and test CER after fine-tuning with B1C1G1M1 augmentation on 16 and 64 adaptation lines on all target writer datasets (the writer ID in the header corresponds to the ID in Figure~\ref{fig:dataset:source_and_target}).}\label{tab:target_absolute_results}
\centering
{
\begin{tabular}{c | p{4.5mm} | p{4.5mm} | p{4.5mm} | p{4.5mm} | p{6mm} | p{4.5mm} | p{4.5mm} | p{6mm} | p{4.5mm} | p{4.5mm} | p{6mm} | p{4.5mm} | p{4.5mm} | p{4.5mm} | p{4.5mm} | p{6mm} | p{4.5mm} | p{4.5mm} | p{4.5mm} }
 & 0 & 1 & 2 & 3 & 4 & 5 & 6 & 7 & 8 & 9 & 10 & 11 & 12 & 13 & 14 & 15 & 16 & 17 & 18 \\
\hline
0 & 6.5 & 2.3 & 2.5 & 1.0 & 12.7 & 4.1 & 1.8 & 10.2 & 1.4 & 1.3 & 11.0 & 9.4 & 1.2 & 2.0 & 2.1 & 14.5 & 0.6 & 0.6 & 13.2 \\
16 & 3.3 & 2.6 & 1.3 & 1.0 & 8.3 & 3.7 & 1.6 & 4.9 & 1.2 & 1.0 & 7.4 & 6.8 & 1.0 & 1.4 & 1.2 & 12.9 & 0.4 & 0.6 & 10.1 \\
64 & 2.4 & 2.3 & 1.0 & 1.0 & 6.6 & 3.1 & 1.5 & 2.9 & 0.9 & 0.8 & 5.5 & 5.2 & 0.8 & 1.2 & 1.0 & 11.0 & 0.3 & 0.5 & 7.8 \\
\end{tabular}}
\end{table}
Table~\ref{tab:target_absolute_results} shows the test CER of the baseline model (0) and the test CER after fine-tuning with B1C1G1M1 augmentation on 16 and 32 adaptation lines on all target writer datasets (the writer ID in the header corresponds to the ID in Figure~\ref{fig:dataset:source_and_target}).

\begin{figure*}[t]
    \centering
    \begin{subfigure}[t]{0.5\textwidth}
        \centering
        \includegraphics[width=\textwidth, trim=0mm 0mm 0mm 0mm, clip]{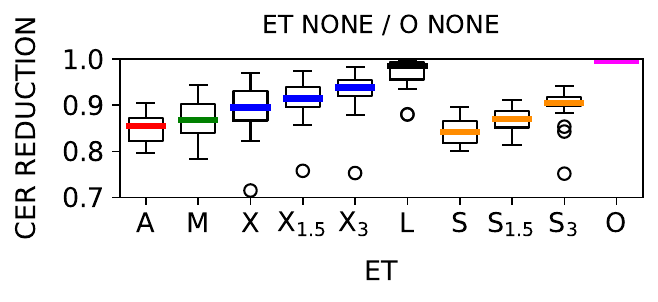}
    \end{subfigure}%
    ~
    \begin{subfigure}[t]{0.5\textwidth}
        \centering
        \includegraphics[width=\textwidth, trim=0mm 0mm 0mm 0mm, clip]{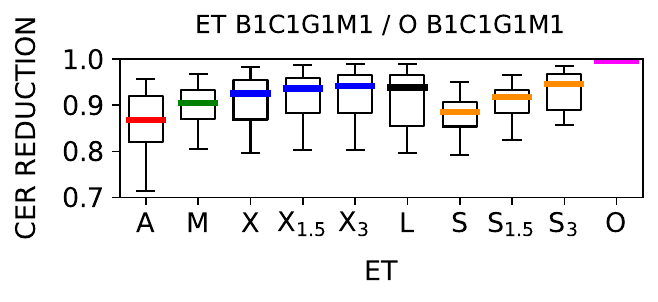}
    \end{subfigure}
    \caption{Compares different estimation strategies (ET) for choosing the fine-tuning iterations with minimal test CER.
    The performance is expressed as a normalized relative reduction of the baseline test CER.}
    \label{fig:writer_finetuning:estimation_techniques}
\end{figure*}
Figure~\ref{fig:writer_finetuning:estimation_techniques}
compares different estimation strategies for choosing the optimal number of fine-tuning iterations.
An estimation strategy is shown as a distribution across the respective normalized writers' CER reductions, where the normalization is done across the estimate strategy dimension with the Oracle strategy, and the line cluster dimension is subsequently aggregated by mean. 
We calculated these statistics only on line clusters 16, 32, 64, 128, and 256, and we omitted three writers, as the normalization was not possible because some fine-tuned models worsen the performance of the baseline.
The best estimation strategy for fine-tuning without adaptation is L.
Generally, the estimation strategy which provides more fine-tuning iterations is better, we give our explanation of this phenomenon while discussing the fine-tuning curves.
For fine-tuning with augmentation $\mathrm{X}_{1.5}$ and $\mathrm{X}_3$ brought the largest CER reductions among cross-validation approaches, and $\mathrm{S}_3$ among the static iteration approaches.

For $\mathrm{S}_3$, the ratios between the estimated number of fine-tuning iterations and the number of adaptation lines were 180, 90, 60, 38, 33, 30, 24, 15, and, 9, for 1, 2, 4, 8, 16, 32, 64, 128, and 256 lines, respectively.
The difference between the ratios of different writers was negligible.
This suggests that there is a fixed relation between the number of optimal fine-tuning iterations and the number of adaptation lines, which on average outperforms the cross-validation approaches.
The target writer distribution for the L strategy is more skewed towards the poorer CER reductions, note that L can be seen as another variant of S.

\begin{figure*}[t]
    \centering
    \begin{subfigure}[t]{0.5\textwidth}
        \centering
        \includegraphics[width=\textwidth, trim=0mm 0mm 0mm 0mm, clip]{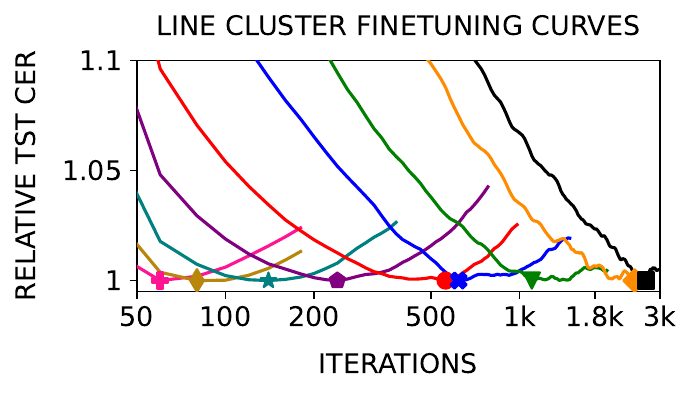}
    \end{subfigure}%
    ~ 
    \begin{subfigure}[t]{0.5\textwidth}
        \centering
        \includegraphics[width=\textwidth, trim=0mm 0mm 0mm 0mm, clip]{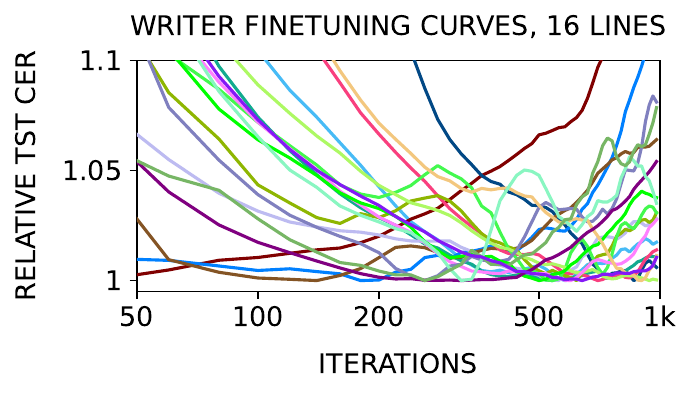}
    \end{subfigure}

    \begin{subfigure}[t]{0.5\textwidth}
        \centering
        \includegraphics[width=\textwidth, trim=0mm 0mm 0mm 0mm, clip]{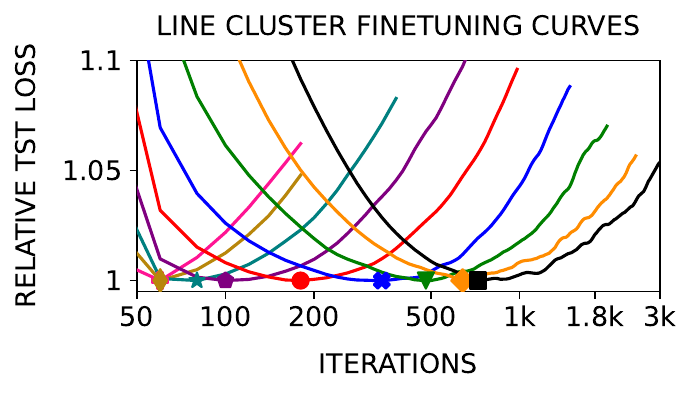}
    \end{subfigure}%
    ~ 
    \begin{subfigure}[t]{0.5\textwidth}
        \centering
        \includegraphics[width=\textwidth, trim=0mm 0mm 0mm 0mm, clip]{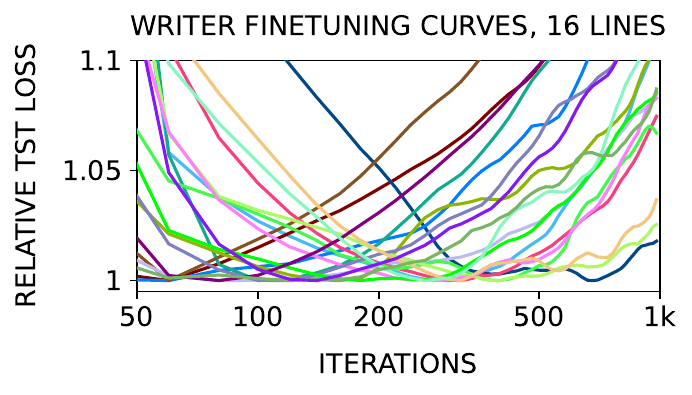}
    \end{subfigure}

    \begin{subfigure}[t]{\textwidth}
        \centering
        \includegraphics[width=\textwidth, trim=0mm 0mm 0mm 85mm, clip]{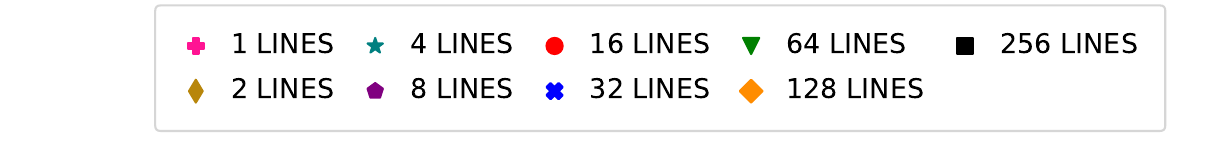}
    \end{subfigure}
    \caption{Fine-tuning curves for B1C1G1M1 augmentation.}
    \label{fig:finetuning_curves}
\end{figure*}
\subsubsection{Fine-tuning curves.}
To get a deeper insight into the fine-tuning process with B1C1G1M1 augmentation, we show aggregations of fine-tuning curves in Figure~\ref{fig:finetuning_curves}.
Line cluster fine-tuning curves for each line cluster in the left column graphs were computed with the Static Iteration (S) estimation strategy on all 19 target writers, for the CER graphs, the calculation is based on the test CER fine-tuning curves.
The graphs in the right column show writer fine-tuning curves for line cluster 16 before aggregation, note that the colors match the colors in Figure~\ref{fig:dataset:source_and_target} and Figure~\ref{fig:writer_finetuning}.    

On average, for all line clusters, the fine-tuning curves had a U-like shape and the minimum test CER was always achieved later than the respective minimum test loss.
For line cluster 16, the amount of fine-tuning iterations to achieve the optimal CER reduction varied among different writers, and some of them (darker brown and blue) suffered from overtraining, this can also be seen in Figure~\ref{fig:writer_finetuning}.
The CER fine-tuning curves were smooth and had a negative slope up until the loss curves started to grow more dramatically, from this point they were prone to high noise.
This phenomenon is more drastic for cross-validation, especially for a lower amount of adaptation lines.
Therefore, the estimation of the minimal test CER fine-tuning iteration based on loss fine-tuning curves should be derived from iterations of a slightly uncertain region behind the minimum.
Returning to the left graphs, we can see that on average the optimal level of uncertainty is higher for higher amounts of adaptation lines, which is the motivation behind $\mathrm{X_R}$ and $\mathrm{S_R}$ estimation strategies.

By inspecting the fine-tuning curves for the fine-tuning without augmentation, we found out that the baseline model quickly overfitted the adaptation lines.
The loss on test lines got to the minimum around the first 100 iterations and started to increase afterward.
Surprisingly, at this point, the CER saturated or even kept getting slightly better for the remaining iterations.
This phenomenon might have been caused by the fact that after the model overfitted the adaptation lines the training loss was minimal and the subsequent iterations produced only slightly less confident models which turned out to be more accurate.
This explains why using the L estimation strategy brought the best CER reductions.

\subsection{Writer-dependent Scenario}

\begin{figure*}[t]
    \centering
    \begin{subfigure}[t]{0.5\textwidth}
        \centering
        \includegraphics[width=\textwidth, trim=0mm 0mm 0mm 0mm, clip]{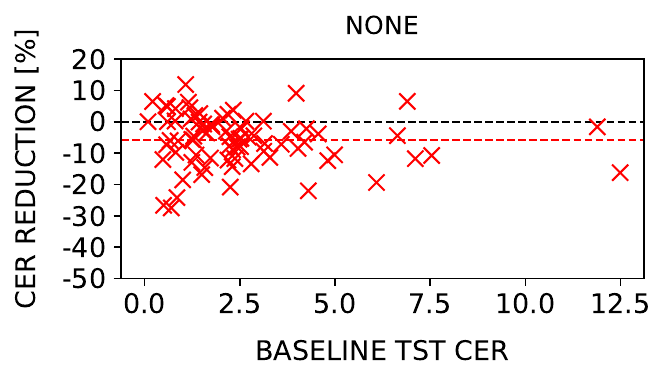}
    \end{subfigure}%
    ~
    \begin{subfigure}[t]{0.5\textwidth}
        \centering
        \includegraphics[width=\textwidth, trim=0mm 0mm 0mm 0mm, clip]{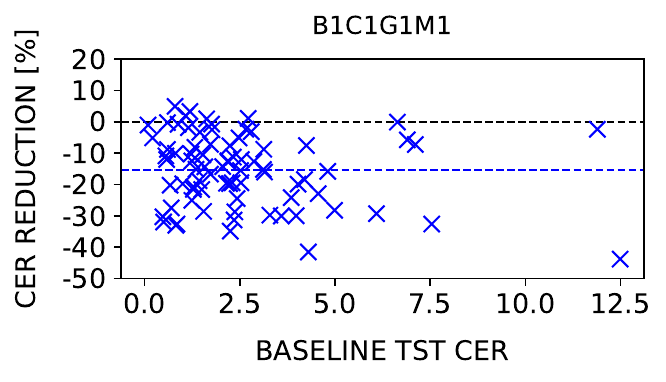}
    \end{subfigure}
    \caption{Relative test CER reductions for fine-tuning in the writer-dependent scenario on 78 writers from the CzechHWR dataset together with the baseline test CER.}
    \label{fig:writer_dependent_finetuning}
\end{figure*}

This section describes fine-tuning of the baseline model on writers from the source CzechHWR dataset.
To cover different numbers of training lines, we chose one random writer per each group of writers with the same number of lines, which resulted in 315 writers.
The baseline model was fine-tuned for 1000, 2000, 3000, and 6000 iterations for writers with the number of lines more than or equal to 1, 100, 500, and 1000.
To eliminate noise bias from the result statistics, we estimated a function that took the number of writer training lines as the input and output the number of fine-tuning iterations.
The estimation was done as a polynomial fitting on a dataset of $(N, I)_W$ tuples, where $W$ was the fine-tuned writer, $N$ was the number of its training lines, and $I$ was the fine-tuning iteration with minimal test CER.
Polynomial fitting with additional parameters in the form of train and test loss/CER did not bring any improvements.    

Figure~\ref{fig:writer_dependent_finetuning} shows the relative  CER reductions for 78 writers together with the baseline test CER.
We do not show results for writers with less than 500 training lines, due to the insufficient number of testing lines in the CzechHWR dataset (see Table~\ref{tab:dataset}).
The colored dashed lines are the means of the writers' CER reductions.
Fine-tuning without augmentation was prone to overfitting but still brought a 6\% CER reduction on average.
Fine-tuning with B1C1G1M1 augmentation almost eliminated overfitting and brought a 15\% CER reduction on average.
These results show that our baseline model was not able to handle a vast number of writing styles present in the CzechHWR dataset, even though it was well-trained and for the last 100k iterations with a small learning rate did not bring any further improvements.
We believe that fine-tuning in this writer-dependent scenario allows the model to adapt to otherwise ambiguous aspects of the text -- that it is not just due to a low modeling capacity of the model with respect to the size and variability of the general dataset.
An ensemble of writer-dedicated models, where each of these models would be a fine-tuned variant of the shared baseline model, seems to be a reasonable baseline for handwritten text recognition in the writer-dependent scenario.

\section{Conclusion}

Our experiments show that fine-tuning is a very efficient domain adaptation method for handwritten text recognition.
In the writer-independent scenario, it improved the recognition accuracy of the baseline model by 20\,\% to 45\,\% relatively, for 16--256 adaptation lines, when choosing the number of fine-tuning iterations by cross-validation.
We further showed that it is possible to estimate a fixed ratio between the number of fine-tuning iterations and the number of adaptation text lines, which outperformed the cross-validation technique.
This indicates that in live handwriting recognition applications, this mapping can be estimated for a specific general model on a small number of exemplar documents and that fine-tunning for new documents can be performed with a predefined number of iterations conditioned only on the amount of available target data without risking overfitting or accuracy degradation. 
This fine-tuning with this fixed stopping criterion works even for a very small number of text lines. 
In our experiments, the improvements for 2--8 text lines were 10\% to 20\% on average, and even a single  adaption text line without augmentations improved transcription accuracy by 5\,\% on average.
Fine-tuning was surprisingly resistant to overfitting even for an extremely low number of text lines and the region of an optimal number of fine-tuning iterations proved to be wide and easy to localize. 
Data augmentation proved to be an important component of the fine-tuning process with a combination of geometry, blur, and noise masking providing $1.5\times$ larger improvement over fine-tuning without any augmentation.

Surprisingly, the fine-tuning was effective also on documents from the original training set (in the writer-dependent scenario) where the observed improvement reached 15\,\%. 

The experimental result reported in this paper has strong practical implications for handwriting recognition applications. 
The conclusion is that this type of fine-tuning should be always used and that it is safe to do so. 
We have already implemented this strategy in our text recognition web application PERO OCR\footnote{https://pero-ocr.fit.vutbr.cz}, where users can repeatedly transcribe a document, where each transcription first fine-tunes the selected model to already corrected lines in the document.

We have performed preliminary experiments with Transformer-based sequence-to-sequence models.
They tend to overfit the adaptation text lines while the noise masking augmentation makes the overfitting even worse. 
We presume that such behavior is due to the autoregressive decoder which learns the text of the adaptation lines. 
We are looking at several methods how to mitigate this behavior including self-training with dedicated language models, constraining the change of the model, and others.

\subsubsection*{Acknowledgment.} This work has been supported by the Ministry of Culture Czech Republic in NAKI III project semANT - Semantic Document Exploration (DH23P03OVV060).

%
%

\bibliographystyle{splncs04}
\bibliography{mybibliography}

\end{document}